\documentclass[letterpaper,10pt,conference]{ieeeconf}

\IEEEoverridecommandlockouts{}
\usepackage{amsmath}
\usepackage{amssymb}
\usepackage{graphicx}
\usepackage{booktabs}
\usepackage{tabularx}
\usepackage{array}
\usepackage{cite}
\usepackage{balance}

\newcolumntype{Y}{>{\raggedright\arraybackslash}X}
\newcommand{\ind}{\mathbb{I}}
\newcommand{\powermodel}{history-aware power estimator}

\renewcommand{\arraystretch}{1.0}
\AtBeginDocument{%
    \setlength{\abovedisplayskip}{4pt plus 2pt minus 2pt}%
    \setlength{\belowdisplayskip}{4pt plus 2pt minus 2pt}%
    \setlength{\abovedisplayshortskip}{2pt plus 1pt minus 1pt}%
    \setlength{\belowdisplayshortskip}{2pt plus 1pt minus 1pt}%
}

\title{\LARGE \bf
Learning to Exploit Passive Dynamics for Energy-Efficient Target Hopping of a Spring-Legged Quadcopter}

\author{Ruigang Chen$^{1,2}$, Qi Zhang$^{1}$, Zhicheng Zhong$^{1}$, Zhuorui Yun$^{1}$, Yizhar Or$^{2}$, and Mingyi Liu$^{1,2}$
\thanks{*This work was supported by the startup fund from Guangdong Technion – Israel Institute of Technology (Corresponding author: Mingyi Liu).}%
\thanks{This work has been submitted to the IEEE for possible publication. Copyright may be transferred without notice, after which this version may no longer be accessible.}
\thanks{$^{1}$Department of Mechanical Engineering and Robotics, Guangdong Technion - Israel Institute of Technology, Shantou, 515063, Guangdong, China.}%
\thanks{$^{2}$Department of Mechanical Engineering, Technion-Israel Institute of Technology, Haifa, 3200003, Israel.}%
}

\begin{document}

\addtolength{\topmargin}{7.5pt}

\maketitle
\thispagestyle{empty}
\pagestyle{empty}

\begin{abstract}
Combining aerial thrust with spring-loaded hopping makes monopedal quadcopters promising for locomotion over complex terrain, but heuristic proportional--integral--derivative (PID) tuning limits coordination between active thrust and passive contact dynamics. We present a direct estimated-state-to-motor Proximal Policy Optimization (PPO) policy that commands four motors without an explicit hopping state machine or low-level attitude PID. Its reward combines Energy-Manifold Shaping for mass-normalized vertical-energy tracking and apex-state anchoring with Efficiency Shaping, which uses a history-aware power estimator to penalize general power use, impose an additional airborne-power cost, and penalize airborne near-stationarity. In representative hardware runs, the PPO-based control stack reduced cycle-averaged measured electrical power by 30.7\% and mean total normalized thrust by 49.8\% relative to the tuned PID-based control stack, while retaining repeatable commanded-height hopping and more concentrated landings. These observations are consistent with improved use of passive dynamics and reduced measured electrical demand.
\end{abstract}
\section{Introduction}

Biological jumping combines elastic storage, directional control, terrestrial adaptation, and legless mechanisms \cite{richards2017kinematic,gvirsman2016dynamics,jung2017effect,siwanowicz2017three,brunt2016amphibious,ribak2011jumping}. Bioinspired robots demonstrate diverse jumping modes \cite{zhang2017survey,zaitsev2015locust,haldane2017repetitive,yu2016development}, while spring-loaded inverted-pendulum models describe ballistic flight and compliant stance \cite{raibert1986legged,poulakakis2009spring,saranli2010approximate}.

Reviews emphasize integrated design, estimation, actuation, perception, and feedback \cite{siciliano2009robotics,he2020mechanism}. Prior control includes nonlinear model predictive control for a three-dimensional hopper \cite{csomay2023nonlinear}, quadrotor control \cite{bouabdallah2007design}, PogoDrone \cite{zhu2022pogodrone}, and phase-based hopping with a reported 28\% pulse-width modulation (PWM) duty ratio \cite{bai2024agile}; related work studies bidirectional-thruster hopping \cite{li2025high} and jumping--flying trajectories \cite{huang2024real}.

Deep reinforcement learning offers contact-rich control alternatives \cite{arulkumaran2017deep,schulman2017proximal,muzio2022deep,kumagai2025reinforcement}, but objectives can reward unintended surrogates \cite{yuan2019novel,knox2024learning,ibrahim2024comprehensive}. We map a 37-dimensional observation to four motors using Energy-Manifold Shaping and Efficiency Shaping; apex height and horizontal target are commands, with the latter fixed for landing evaluation.

Our contributions are:
\begin{itemize}
    \setlength{\itemsep}{0pt}
    \setlength{\parsep}{0pt}
    \setlength{\topsep}{0pt}
    \setlength{\partopsep}{0pt}
    \item a direct estimated-state-to-motor PPO policy mapping target-conditioned state estimates to four motors without an explicit hopping state machine or attitude PID;
    \item Energy-Manifold Shaping for vertical-energy and apex-state regulation, and Efficiency Shaping for general power, airborne-power, and airborne near-stationarity penalties, supported by a \powermodel{}; and
    \item hardware evaluation of the PPO- and PID-based control stacks for power, thrust, descent, height, landing, and endurance, with reward-module ablations and held-out estimator validation.
\end{itemize}

\begin{figure}[!t]
    \centering
    \includegraphics[width=\linewidth]{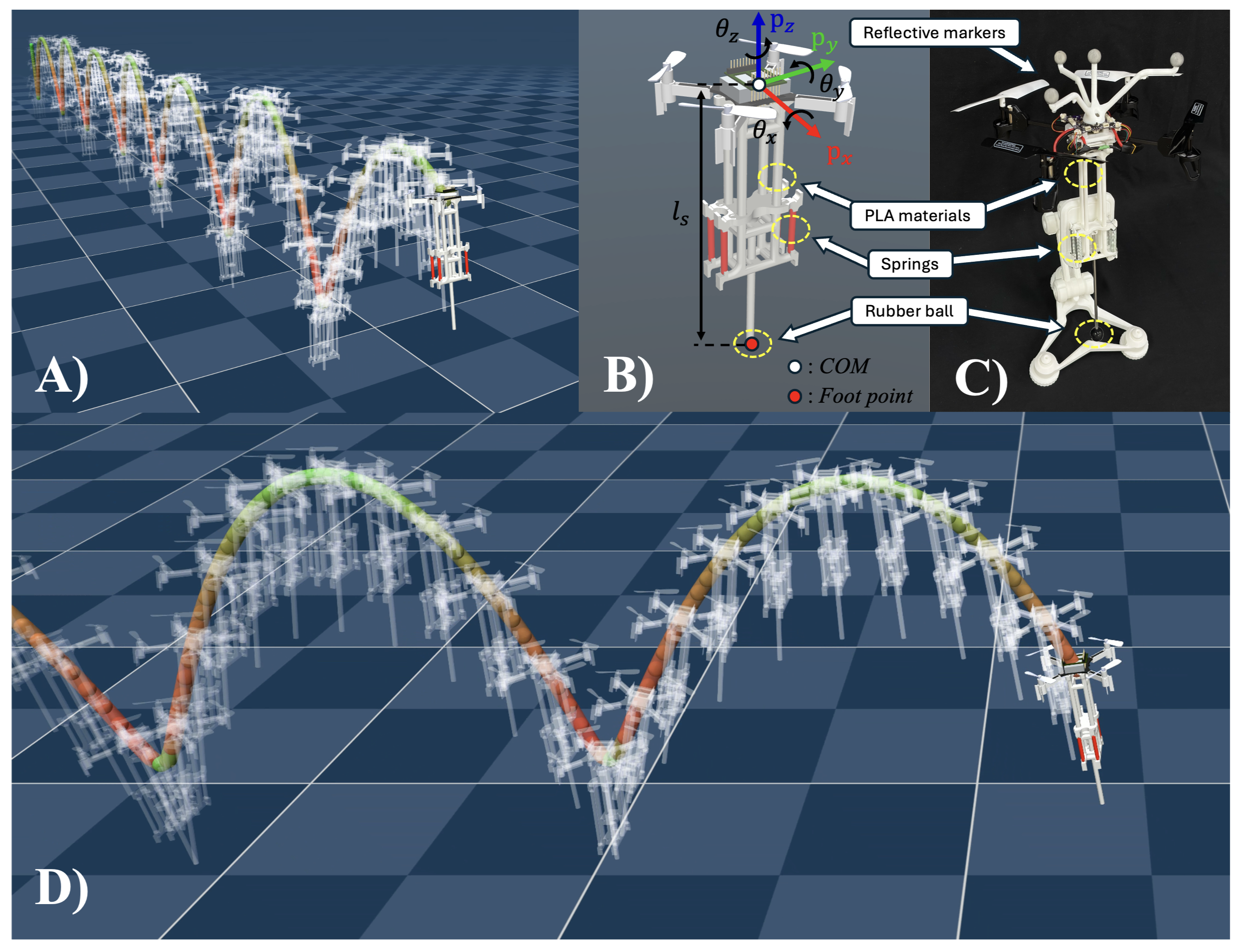}
    \caption{Platform overview: (A) hopping chronophotography, (B) mechanical model and center-of-mass (COM) coordinates, (C) prototype with a 3-D-printed polylactic acid (PLA) leg base, and (D) repeated trajectory. In (B), $p_x,p_y,p_z$ are components of $\mathbf p$; $\theta_x,\theta_y,\theta_z$ visualize roll, pitch, and yaw of quaternion $\mathbf q$; and $l_s$ is the illustrated leg length.}
    \label{fig:platform}
\end{figure}

Fig.~\ref{fig:platform} summarizes the platform and mechanical notation.

\section{Methodology}

\subsection{Hybrid Dynamics and Identified Physical Parameters}

Let $\mathbf p=[p_x,p_y,p_z]^\top$ and $\mathbf v=[v_x,v_y,v_z]^\top$ denote world-frame COM position and velocity, $\mathbf q=[q_w,q_x,q_y,q_z]^\top$ the scalar-first unit orientation quaternion, $\boldsymbol\omega=[\omega_x,\omega_y,\omega_z]^\top$ body angular velocity, and $q_s,\dot q_s$ the passive prismatic spring coordinate and velocity. Increasing $q_s$ corresponds to increasing spring compression and stored elastic potential energy, whereas decreasing $q_s$ corresponds to spring release. No direct spring-position sensor is used on the hardware; instead, $q_s$ is reconstructed deterministically from the known leg geometry and measured body pose and orientation, and $\dot q_s$ is obtained from its temporal evolution. The state is
\begin{equation}
    \mathbf x=
    [\mathbf p^\top,\mathbf v^\top,\mathbf q^\top,
    \boldsymbol\omega^\top,q_s,\dot q_s]^\top.
\end{equation}
The command is $\mathbf p^*=[(\mathbf p_{xy}^*)^\top,z^*]^\top$, where $\mathbf p_{xy}=[p_x,p_y]^\top$, $\mathbf p_{xy}^*=[p_x^*,p_y^*]^\top$, and $z^*$ is the commanded apex height. Let $\mathbf u=[u_1,u_2,u_3,u_4]^\top$ denote the motor-command vector, $c_t\in\{0,1\}$ the contact indicator ($0$ in flight and $1$ in contact), $\mathbf w_c$ the contact wrench, and $f_{\mathrm{air}},f_{\mathrm{contact}}$ the aerial and contact vector fields. The hybrid dynamics are
\begin{equation}
    \dot{\mathbf x}=
    \begin{cases}
        f_{\mathrm{air}}(\mathbf x,\mathbf u), & c_t=0,\\
        f_{\mathrm{contact}}(\mathbf x,\mathbf u,\mathbf w_c),
        & c_t=1,
    \end{cases}
\end{equation}
Here $m$ is total mass, $\mathbf J=\operatorname{diag}(I_{xx},I_{yy},I_{zz})$ body inertia, $\mathbf e_3=[0,0,1]^\top$, $g=9.81~\mathrm{m\,s^{-2}}$ gravitational acceleration, and $\mathbf g=[0,0,-g]^\top$ world-frame gravity. The matrix $R(\mathbf q)$ maps body-frame vectors to the world frame; $F_z$ is resultant rotor thrust along body $\mathbf e_3$; $\boldsymbol\tau=[\tau_x,\tau_y,\tau_z]^\top$ is rotor moment; and $\mathbf w_c=[\mathbf f_c^\top,\boldsymbol\tau_c^\top]^\top$ comprises contact force and moment. The rigid-body translational and rotational subsystem is
\begin{equation}
\left\{
\begin{array}{@{}l@{\;}l@{}}
    \dot{\mathbf p}
      &={}\mathbf v,\\
    m\dot{\mathbf v}
      &={}m\mathbf g+R(\mathbf q)\mathbf e_3F_z+\mathbf f_c,\\
    \mathbf J\dot{\boldsymbol\omega}
      +\boldsymbol\omega\!\times\!(\mathbf J\boldsymbol\omega)
      &={}\boldsymbol\tau+\boldsymbol\tau_c.
\end{array}
\right.
\end{equation}
Quaternion kinematics and passive prismatic-joint evolution are advanced numerically with the PhysX contact constraints. The leg-transmitted contact wrench vanishes in flight; stance may be closed analytically using unilateral/friction constraints and an impact law \cite{goebel2009hybrid,brogliatononsmooth}, or numerically as described next. The spring coordinate $q_s=0$ is the zero-preload reference. With measured $k_s=604$~N/m and $d_s=2$~N\,s/m, the passive spring force is
\begin{equation}
    F_s=-k_sq_s-d_s\dot q_s,
\end{equation}
This restoring force contributes to the contact-mediated dynamics represented by $\mathbf w_c$.

Table~\ref{tab:physical} lists hardware and simulation parameters. Principal inertias were identified by a diagonal least-squares rigid-body fit to reconstructed motor moments and gyroscope derivatives \cite{eschmann2024sysid}. The 0.183~kg platform uses four 3.7-V 8520 motors, 14-cm propellers, AO4466 drives, extension springs, and printed PLA structure.

\begin{table}[!t]
\caption{Core Physical and Simulation Parameters}
\label{tab:physical}
\centering
\scriptsize
\setlength{\tabcolsep}{3pt}
\begin{tabularx}{\columnwidth}{@{}YY@{}}
\toprule
Parameter & Value \\
\midrule
Total measured mass & 0.183~kg \\
Body/SpringLeg simulation masses & 0.173/0.010~kg \\
Motors/drive stage & $4\times$ 3.7-V 8520 / AO4466 MOSFET \\
Propeller length & 0.14~m \\
Spring type & Stainless-steel extension spring \\
$3$-D printed structure & Bambu Lab PLA / A1 printer \\
$I_{xx}$ & $1.23\times10^{-3}$~kg\,m$^2$ \\
$I_{yy}$ & $1.29\times10^{-3}$~kg\,m$^2$ \\
$I_{zz}$ & $2.31\times10^{-3}$~kg\,m$^2$ \\
SpringLeg diagonal inertia & $1.0\times10^{-5}$~kg\,m$^2$ \\
Quadrotor diagonal dimension & 0.230~m \\
Motor offset $L$ & 0.0813~m \\
Spring-joint axis/range & Body $Z$ / 0--0.08~m \\
Spring stiffness/damping & 604~N/m / 2~N\,s/m \\
Spring-joint numerical force/velocity limits & 200~N / 12~m/s \\
Static/dynamic friction & 1.5/1.5 \\
Physics and control rate & 100~Hz \\
\bottomrule
\end{tabularx}
\vspace{1pt}
\parbox{\columnwidth}{\scriptsize MOSFET: metal--oxide--semiconductor field-effect transistor.}
\end{table}

\subsection{Isaac Lab Simulation and Passive-Leg Model}

Isaac Lab uses NVIDIA PhysX for collisions, constraints, and impact impulses \cite{nvidia2024physx}. The Universal Scene Description asset has a main body and passive prismatic SpringLeg with $q_s\in[0,0.08]$~m along body $Z$, zero preload, and zero restitution. Physics and control run at 100~Hz for 15-s episodes in 4096 environments.

\subsection{PPO Command Interface, Motor Dynamics, and Wrench Mapping}

PPO uses an actor to map observations to actions and a critic during optimization \cite{schulman2017proximal}; the deployed actor directly commands four motors. The operator $\operatorname{clip}(x,a,b)$ denotes elementwise clipping to $[a,b]$. At step $t$, the actor output $\mathbf a_t=[a_{1,t},\ldots,a_{4,t}]^\top\in[-1,1]^4$ is mapped to
\begin{equation}
    \mathbf u_{\mathrm{target},t}
    =\operatorname{clip}(0.5\mathbf a_t+0.5,0,1).
\end{equation}
The endpoints represent 0\% and 100\% PWM. Because $\mathrm{PWM}_i=1000u_i$, $u_i=1$ is the full 1000-count hardware range; clipping remains a deployment safeguard against numerical or model-export deviations.

With control interval $\Delta t=0.01$~s and motor time constant $\tau_m$, define $\alpha=\Delta t/(\tau_m+\Delta t)$. Training samples delay $d\in\{2,3,4\}$ steps (20--40~ms), while testing uses $d=3$. The realized command $\mathbf u_t=[u_{1,t},\ldots,u_{4,t}]^\top$ obeys
\begin{equation}
    \mathbf u_t=\alpha\mathbf u_{\mathrm{target},t-d}
      +(1-\alpha)\mathbf u_{t-1},
\end{equation}
where $\mathbf u_{t-1}$ is its previous value; $\tau_m$ is sampled over $[0.10,0.14]$~s and fixed at 0.125~s for testing.

Suppressing the time index, let $F_i$ be motor $i$'s thrust and $u_i$ its realized command. Fitting motor--propeller data gives
\begin{equation}
    F_i=-0.237u_i^2+0.813u_i+0.0113,
    \label{eq:thrust_mapping}
\end{equation}
The constant term in (\ref{eq:thrust_mapping}) is an unconstrained empirical regression offset over the identified operating data. It is retained as part of the fitted mapping but is not interpreted as physical thrust produced by a de-energized motor. Accordingly, (\ref{eq:thrust_mapping}) should be understood as an empirical approximation over the identified motor-command range rather than a first-principles motor model. Here $F_i$ is in newtons along body $\mathbf e_3$. With $L=0.0813$~m and $K_\tau=5.4\times10^{-2}$~N\,m, the rotor wrench is
\begin{equation}
\left\{
\begin{aligned}
    F_z
      &=F_1+F_2+F_3+F_4,\\
    \tau_x
      &=L(F_1+F_4-F_2-F_3),\\
    \tau_y
      &=L(F_1+F_2-F_3-F_4),\\
    \tau_z
      &=-K_\tau(u_1^2+u_3^2-u_2^2-u_4^2).
\end{aligned}
\right.
\end{equation}
Motors 1 and 3 counter-rotate relative to motors 2 and 4, producing the signed reaction-torque difference in $\tau_z$. Here $\mathcal U[a,b]$ denotes the uniform distribution on $[a,b]$. Training samples $\eta_F\sim\mathcal U[0.95,1.05]$ and $\eta_{\tau,x},\eta_{\tau,y},\eta_{\tau,z}\sim\mathcal U[0.9,1.1]$, applying $F_z/\eta_F$ and $\tau_j/\eta_{\tau,j}$ for $j\in\{x,y,z\}$ without changing mass or inertia.

\subsection{Observation and Action Spaces}
\label{sec:observation_space}

\noindent
\resizebox{\columnwidth}{!}{$\displaystyle
\begin{aligned}
\mathbf o_t&=\big[({}^{B}\mathbf v_t)^\top,({}^{B}\boldsymbol\omega_t)^\top,\mathbf q_t^\top,
(\delta\mathbf p_t^{B})^\top,z_t,c_t,q_{s,t},\dot q_{s,t},\mathbf h_{a,t}^\top\big]^\top\in\mathbb R^{37},\\
\mathbf h_{a,t}&=\big[\mathbf a_{t-5}^\top,\mathbf a_{t-4}^\top,\mathbf a_{t-3}^\top,
\mathbf a_{t-2}^\top,\mathbf a_{t-1}^\top\big]^\top\in\mathbb R^{20}.
\end{aligned}$}
\par
Here $^{B}\mathbf v_t,{}^{B}\boldsymbol\omega_t\in\mathbb R^3$ are body-frame linear/angular velocities, $\mathbf q_t\in\mathbb R^4$ is scalar-first, and $\delta\mathbf p_t^{B}=R(\mathbf q_t)^\top(\mathbf p^*-\mathbf p_t)\in\mathbb R^3$ is body-frame target-relative position. The scalars $z_t,c_t,q_{s,t},\dot q_{s,t}$ give COM height, contact, and spring state. The history contains the five preceding actor actions $\mathbf a_{t-5:t-1}$, not realized commands; the dimensions sum as $3+3+4+3+1+1+1+1+20=37$.
In simulation, the previously defined contact indicator is instantiated as
\begin{equation}
    c_t=\ind[q_{s,t}>0.002~\mathrm{m}],
\end{equation}
where $\ind[\cdot]$ denotes the indicator function.
Zero-mean Gaussian observation noise with standard deviation 0.01 is applied in simulation.

\subsection{PPO Architecture and Training}

Training uses the Robotic Systems Lab Reinforcement Learning (RSL-RL) library \cite{schwarke2025rslrl}. The actor is $\pi_\theta(\mathbf a_t\mid\mathbf o_t)$ and critic $V_\phi(\mathbf o_t)$; define $V_t:=V_\phi(\mathbf o_t)$ and store $\log\pi_\theta(\mathbf a_t\mid\mathbf o_t)$. Feed-forward multilayer perceptrons (MLPs) use 256--128 actor and 256--256 critic layers with exponential linear unit (ELU) activations \cite{clevert2016elu}. The policy receives $\mathbf h_{a,t}$ and uses generalized advantage estimation (GAE) with $\lambda=0.95$; Table~\ref{tab:training} summarizes training.

\begin{table}[!t]
\caption{Core PPO Training Parameters}
\label{tab:training}
\centering
\scriptsize
\setlength{\tabcolsep}{3pt}
\begin{tabularx}{\columnwidth}{@{}Yl@{}}
\toprule
Parameter & Value \\
\midrule
Observation/action dimensions & 37 / 4 \\
Action history & 5 steps (50~ms) \\
Policy class & Feed-forward MLP \\
Actor/critic hidden layers & 256--128 / 256--256 \\
Activation/action-noise standard deviation & ELU / 0.4 \\
Parallel environments/rollout & 4096 / 256 steps \\
Iterations/transitions & 250 / 262,144,000 \\
Learning rate/schedule & $3.0\times10^{-4}$ / adaptive \\
Epochs/mini-batches & 5 / 8 \\
Discount/GAE & $\gamma=0.99$, $\lambda=0.95$ \\
PPO clip/entropy coefficient & 0.2 / 0.01 \\
Target-height/initial-drop range & 1.0--1.5 / 0.8--1.5~m \\
Action delay/motor time constant & 2--4 steps / 0.10--0.14~s \\
Observation-noise standard deviation & 0.01 \\
\bottomrule
\end{tabularx}
\end{table}

\subsection{Energy-Manifold Shaping}

Instantaneous height tracking cannot distinguish a ballistic apex from rotor-supported hovering, so Energy-Manifold Shaping combines vertical-energy tracking and apex-state anchoring. Let $z_t:=p_{z,t}$ and $v_{z,t}$ be COM vertical position and velocity; $\varepsilon_{z,t}$ and $\varepsilon_z^*$ are specific vertical mechanical energy and its target:
\begin{align}
    \varepsilon_{z,t}
      &=g\max(z_t,0)+\frac{1}{2}v_{z,t}^{2},\\
    \varepsilon_z^*&=gz^*.
\end{align}
Because mass is omitted, both have units $\mathrm{m^2\,s^{-2}}$ ($\mathrm{J\,kg^{-1}}$). In generic coordinates $(z,v_z)$, the target level set is
\begin{equation}
    \mathcal M_E=\left\{(z,v_z):
    g\max(z,0)+\frac{1}{2}v_z^2=gz^*\right\}.
\end{equation}
Here ``manifold'' denotes this specific-energy level set, not a formal hybrid-orbit stability guarantee. With tuned $w_E=12$ and $\sigma_E=2.0~\mathrm{m^2\,s^{-2}}$, the energy reward is
\begin{equation}
    r_{E,t}=w_E
    \exp\left(
    -\frac{|\varepsilon_{z,t}-\varepsilon_z^*|}{\sigma_E}
    \right),
\end{equation}
With tuned $w_A=8$, $\sigma_z=0.18$~m, and $\sigma_v=0.35~\mathrm{m\,s^{-1}}$, apex-state anchoring is
\begin{equation}
    r_{A,t}=w_A
    \exp\left(-\frac{|z_t-z^*|}{\sigma_z}\right)
    \exp\left(-\frac{|v_{z,t}|}{\sigma_v}\right),
\end{equation}
The two terms favor the target energy level set and an apex near $(z^*,0)$.

\subsection{History-Aware Power Estimator}

Electrical-drive dynamics motivate recent-actuation history \cite{krause2002analysis}. The \powermodel{} was trained on all 36 ordered transitions among PWM nodes $\{0,100,200,400,600,800\}$ with approximately 1.5-s holds and common/per-motor perturbations of $\pm30/\pm20$ counts. Commands, voltage, and current were logged near 50~Hz, resampled at 20~ms, and normalized to $[0,1]$. For battery voltage $V_{\mathrm{bat},t}$ and current $I_t$, measured electrical power is
\begin{equation}
    P_{\mathrm{elec},t}=V_{\mathrm{bat},t}I_t.
    \label{eq:measured_power}
\end{equation}
Let $\mathbf u_{t-9:t}$ be the ordered ten-step command history and $\bar{\mathbf u}_{15,t},\bar{\mathbf u}_{25,t}$ the 15- and 25-step per-motor averages. Initialize $s_0=0$ and update $s_t=s_{t-1}+10^{-4}\sum_{i=1}^{4}u_{i,t}$ for $t\geq1$. With $\operatorname{vec}(\cdot)$ denoting temporal flattening, the input is
\begin{equation}
    \mathbf z_t^P=[\operatorname{vec}(\mathbf u_{t-9:t})^\top,
    \bar{\mathbf u}_{15,t}^\top,\bar{\mathbf u}_{25,t}^\top,s_t]^\top
    \in\mathbb R^{49},
\end{equation}
Its dimension is $10\times4+4+4+1=49$. For frozen parameters $\xi$, the estimated electrical power is
\begin{equation}
    \widehat P_{\mathrm{elec},t}
    =f_\xi(\mathbf z_t^P),
\end{equation}
in watts. The 0.2/0.3/0.5-s windows feed a 49--128--128--64--32--1 ELU network with Softplus output. Training uses 150 epochs, 512-sample batches, Adam at $10^{-3}$, mean-squared error, and an 85/15 temporal split; the lowest-loss checkpoint is fixed during PPO training.

\subsection{Efficiency Shaping}

Efficiency Shaping is a reward module for reducing electrical demand and powered near-stationarity; it does not estimate electromechanical efficiency as an output-to-input power ratio. Its coefficients are $\lambda_P=0.08$, $\lambda_{\mathrm{air}}=0.6$, and $\lambda_{\mathrm{hov}}=12$, with thresholds $z_{\mathrm{hov}}=0.55$~m and $v_{\mathrm{hov}}=0.25~\mathrm{m\,s^{-1}}$. Using $\widehat P_{\mathrm{elec},t}$:
\begin{align}
r_{\mathrm{eff},t}={}&
-\lambda_P\widehat P_{\mathrm{elec},t} \nonumber\\
&-\lambda_{\mathrm{air}}(1-c_t)
\widehat P_{\mathrm{elec},t} \nonumber\\
&-\lambda_{\mathrm{hov}}(1-c_t)
\ind[z_t>z_{\mathrm{hov}}]
\ind[|v_{z,t}|<v_{\mathrm{hov}}],
\end{align}
The terms penalize general estimated power, airborne power, and airborne near-stationarity, respectively; the last depends on flight state, height, and vertical velocity rather than the power estimate or commands.

\subsection{Baseline Task and Stability Rewards}

The baseline rewards regulate target error, attitude, motion, action smoothness, low-height behavior, survival, and termination. The near-ground attitude term is
\begin{equation}
    r_{\mathrm{att},t}
    =(q_{x,t}^{2}+q_{y,t}^{2})
    \left[1+5\,\operatorname{clip}(1-z_t,0,1)\right].
\end{equation}
Let $d_{xy,t}=\|\mathbf p_{xy,t}-\mathbf p_{xy}^{*}\|_2$ be the horizontal target distance, $\psi_t$ the wrapped yaw error, $\Delta\mathbf a_t=\mathbf a_t-\mathbf a_{t-1}$ the action increment, and $\Delta d_{xy,t}=d_{xy,t-1}-d_{xy,t}$ the one-step target-distance progress. Table~\ref{tab:baseline_reward_terms} defines the remaining terms and their scales or activation thresholds before they enter the weighted reward.

\begin{center}
\begin{minipage}{\columnwidth}
\centering
\refstepcounter{table}\label{tab:baseline_reward_terms}
{\footnotesize TABLE~\Roman{table}\\[-2pt]
\textsc{Baseline Reward-Term Definitions and Thresholds}}\par
\vspace{2pt}
{\scriptsize
\renewcommand{\arraystretch}{0.92}
\setlength{\tabcolsep}{2pt}
\begin{tabularx}{\columnwidth}{@{}>{$}l<{$}>{$}l<{$}Y@{}}
\toprule
\text{Term} & \text{Expression} & Scale or activation threshold \\
\midrule
r_{\mathrm{term}} & \ind\!\left[\substack{z_t<0.05~\mathrm{m}\ \\ \text{or}\ q_{x,t}^2+q_{y,t}^2>0.5}\right] & Either condition \\
r_{\mathrm{surv}} & 1-r_{\mathrm{term}} & Active until termination \\
r_{\mathrm{yaw}} & \psi_t^2 & Continuous \\
r_{d_{xy}} & e^{-d_{xy,t}/0.5} & 0.5~m decay scale \\
r_{d_z} & e^{-|z^*-z_t|/0.4} & 0.4~m decay scale \\
r_{\omega} & \|\boldsymbol\omega_t\|_2^2 & Continuous \\
r_{\mathrm{lat}} & v_{x,t}^2+v_{y,t}^2 & Continuous \\
r_{\mathrm{lazy}} & \ind[z_t<0.30~\mathrm{m}] & 0.30~m height \\
r_{\mathrm{goal}} & \ind[d_{xy,t}<0.35~\mathrm{m}] & 0.35~m target radius \\
r_{xy,\mathrm{err}} & \min(d_{xy,t},1) & Capped at 1~m \\
r_{\mathrm{rate}} & \|\Delta\mathbf a_t\|_2^2 & Continuous \\
r_{\mathrm{progress}} & \operatorname{clip}(\Delta d_{xy,t},-1,1) & Clipped to $[-1,1]$ \\
\bottomrule
\end{tabularx}
}
\end{minipage}
\end{center}

The baseline weighted sum is
\begin{align}
r_{\mathrm{base},t}={}&
5r_{\mathrm{surv}}+15r_{d_{xy}}+2r_{d_z}
-7r_{\mathrm{yaw}}-20r_{\mathrm{att},t} \nonumber\\
&-0.5r_{\omega}-10r_{\mathrm{lazy}}
-1.5r_{xy,\mathrm{err}}+8r_{\mathrm{goal}} \nonumber\\
&-10r_{\mathrm{lat}}-0.5r_{\mathrm{rate}}.
\end{align}
The complete implemented reward is
\begin{align}
R_t={}&\Delta t\left(
r_{\mathrm{base},t}+r_{E,t}+r_{A,t}+r_{\mathrm{eff},t}
\right) \nonumber\\
&+20r_{\mathrm{progress}}-10r_{\mathrm{term}}.
\end{align}

\begin{figure*}[!t]
    \centering
    \includegraphics[width=1\textwidth]{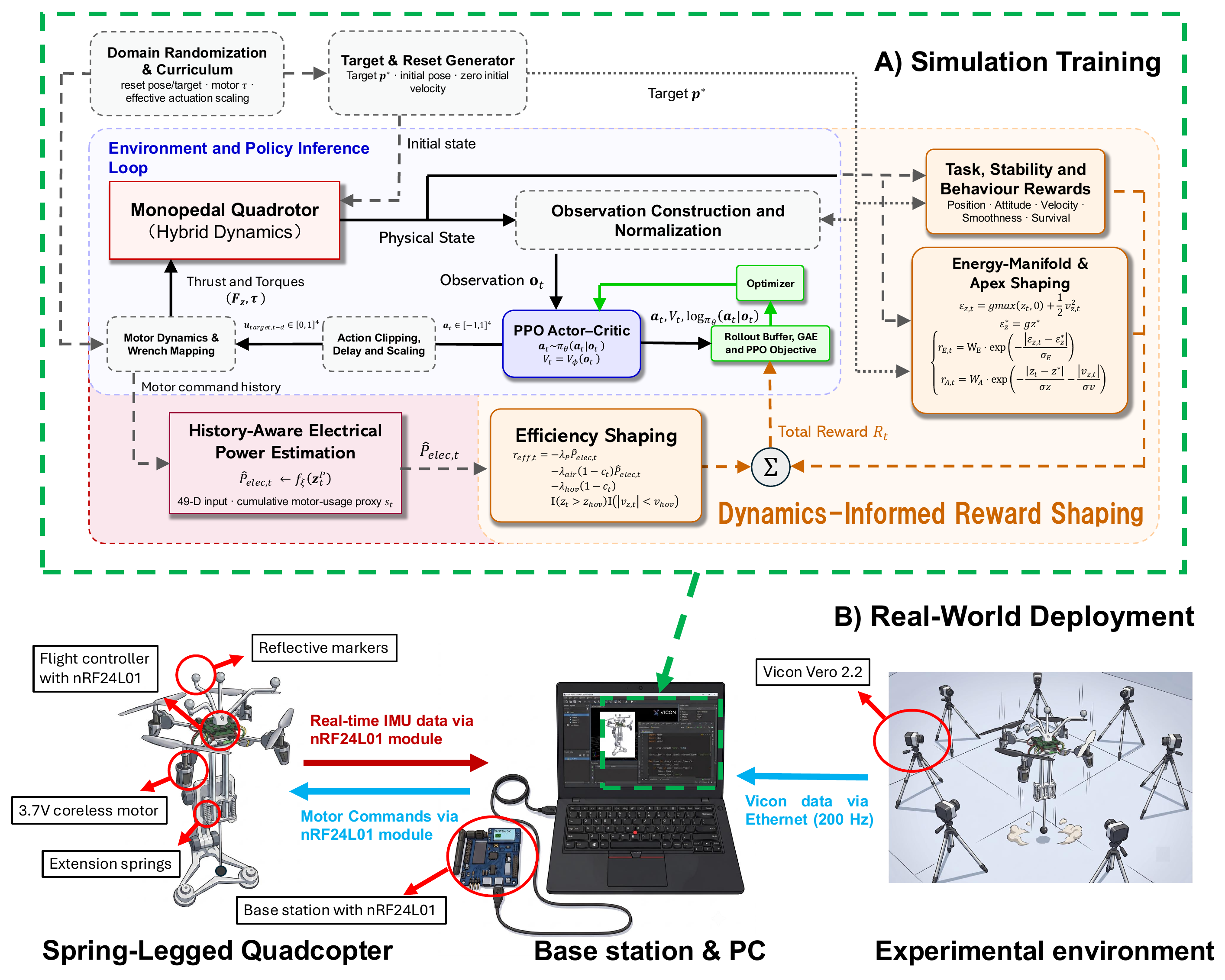}
    \caption{Training and deployment architecture. The PPO actor--critic maps observations through motor dynamics to body wrenches; the history-aware power estimator supplies estimated instantaneous electrical cost to Efficiency Shaping, and Energy-Manifold Shaping regulates vertical energy and apex state. Deployment combines Vicon and inertial sensing, motor communication, and computer-based policy evaluation.}
    \label{fig:pipeline}
\end{figure*}

Fig.~\ref{fig:pipeline} summarizes the resulting training and deployment architecture.

\subsection{Reward-Module and Estimator-Level Ablations}

The ablations isolate the two reward-shaping modules and the history-aware power estimator:
\begin{itemize}
    \setlength{\itemsep}{0pt}
    \setlength{\parsep}{0pt}
    \item \textbf{Full:} uses both shaping modules and the history-aware power estimator.
    \item \textbf{No Energy:} sets $w_E=w_A=0$.
    \item \textbf{No Efficiency:} sets $\lambda_P=\lambda_{\mathrm{air}}=\lambda_{\mathrm{hov}}=0$.
    \item \textbf{Power Proxy:} preserves the reward structure and coefficients but replaces the history-aware estimate with
    \begin{equation}
        \widehat P_{\mathrm{proxy},t}
        =k\sum_{i=1}^{4}u_{i,t}^{2}.
    \end{equation}
\end{itemize}
{\scriptsize
\setlength{\abovedisplayskip}{1pt}
\setlength{\belowdisplayskip}{1pt}
Here $u_i\in[0,1]$ is a dimensionless normalized realized motor command; hence $\sum_i u_i^2$ is dimensionless and $k$ is in watts. Fixed seed 20260718 generated $N=4096$ synthetic sequences (250 100-Hz control steps each) using independently sampled normalized target motor commands in $[0,1]$, $\tau_m\in[0.10,0.14]$~s, and the first-order dynamics in (6). One terminal sample per sequence entered
\begin{equation*}
    k=
    \frac{N^{-1}\sum_{n=1}^{N}\widehat P_{\mathrm{elec}}^{(n)}}
    {N^{-1}\sum_{n=1}^{N}\sum_{i=1}^{4}(u_i^{(n)})^2}
    =24.7608~\mathrm{W}.
\end{equation*}
Thus $N=4096$, not $4096\times250$ independent samples. This ratio of means matches only the frozen estimator's global average output on the predefined synthetic distribution. It uses no measured-power labels, manual reward tuning, or sample-wise least-squares fitting, and does not match time-wise errors, PWM-region variation, or history-dependent transients. Power Proxy is therefore an estimator-level ablation, not a third reward module.
\par}
A run was classified as sustained target hopping when the policy completed the full 15-s evaluation while continuing repeated, stable hopping about the commanded target without falling. A fall or failure to sustain this repeated hopping behavior for the full evaluation was classified as failure. This criterion is a behavioral pass/fail classification rather than a threshold-based tracking-error metric.

\subsection{Static Estimator Baselines and Validation Protocols}

On a 1.3-m, 1,754-sample record, the static quartic estimator is post hoc scaled by $\beta_{\mathrm{oracle}}=1.35$, selected from all measured-power labels to minimize mean absolute error (MAE). This deliberately optimistic, label-informed calibration is not evaluated as a held-out estimate; it corrects global amplitude by jointly scaling the effective $\lambda_P$ and $\lambda_{\mathrm{air}}$ while leaving $\lambda_{\mathrm{hov}}$ unchanged, but cannot recover history-dependent transient shape.

For the second comparison, let $D\in\{1,2,3,4\}$ denote the maximum polynomial degree, $b_0$ and $b_\ell$ the fitted coefficients shared by all motors, and $\ell$ the polynomial-order summation index. We fit the nested static symmetric-additive family
\begin{equation}
    \widehat P_D(\mathbf u_t)
    =b_0+\sum_{\ell=1}^{D}b_\ell\sum_{i=1}^{4}u_{i,t}^{\ell},
    \qquad D\in\{1,2,3,4\}.
\end{equation}
Shared motor coefficients make the estimate permutation invariant, and increasing $D$ adds only the next power. For each polynomial degree, candidate ridge-regularization strengths were fitted using the first 85\% of the fitting record and selected by MAE on the remaining 15\%. After model selection, the selected model was evaluated once on the separate 1,082-sample held-out record used for the reported MAE, root mean square error (RMSE), and coefficient of determination ($R^2$). The held-out record was not used for coefficient fitting or hyperparameter selection, and its temporally correlated samples are not treated as independent replicates.

\subsection{Curriculum and Domain Randomization}

Curriculum expands reset perturbations while domain randomization varies actuation. Targets span 1.0--1.5~m with 0.8--1.5~m initial drops. Difficulty is zero through 40\% of training, reaches unity at 80\%, and then remains fixed. Full resets use area-uniform offsets up to 5~m, roll/pitch $\pm0.25$~rad, and yaw $\pm\pi$. Force/torque effectiveness follows Sec.~II-C; motor time constant and delay follow Table~\ref{tab:training}, with fixed mass and inertia.

The default play protocol uses target $(0,0,1.3)$~m, initial position $(0,0,1.0)$~m, three-step delay, $\tau_m=0.125$~s, and unit effectiveness. Ablations use target/initial position $(0,0,1.0)$~m, zero velocity, and level attitude.

\subsection{Real-Time Deployment Architecture}

The deployment architecture in Fig.~\ref{fig:pipeline}(B) uses an STM32F103C8T6 microcontroller with an MPU6050 inertial measurement unit (IMU), an nRF24L01 radio link, an eight-camera Vicon system operating at 200~Hz, and a personal computer (PC) evaluating the Open Neural Network Exchange (ONNX)-format policy at 100~Hz. A 15-state error-state Kalman filter fuses bias-corrected IMU and timestamped Vicon pose through buffered delayed-update replay. The resulting pose/orientation, velocity, angular-rate, contact, target-relative, geometry-reconstructed spring-state, and five-step preceding-action quantities form the Sec.~\ref{sec:observation_space} observation. Contact uses $z_t<0.295$~m from observed COM touchdown heights; three unhealthy samples, IMU age above 50~ms, or command/radio faults zero PWM.

\section{Experimental Setup}
\label{sec:experimental_setup}

\subsection{Controllers and Measurement Architecture}

The two control stacks were evaluated in separate runs on the same platform. The PID-based control stack is phase-scheduled and cascaded at 200~Hz; contact uses 0.295/0.291~m Vicon-height hysteresis and zero stance commands. Positive rebound velocity triggers 0.15~s at normalized collective 0.90, then normalized aerial baseline 0.15.

During descent, ballistic prediction estimates touchdown time/position. Horizontal feedback generates target roll/pitch below 0.55~s time-to-contact, with a final phase below 0.06~s and $7^\circ$ limit. Roll/pitch rate gains are $(1.8,0.5,0.05)$, yaw gains $(3.0,0.5,0.05)$, and raw PWM outputs clip to $[0,1000]$.

Landing position/velocity gains are 1.2/0.65, with a 0.05~m deadband, 0.6~m/s speed limit, $5.0^\circ/(\mathrm{m/s})$ velocity-to-tilt factor, outer attitude gain 5.0, and a $\pm200$ raw PWM-count correction limit.

The PID-only estimator uses a 400-sample bias calibration, 40-Hz low-pass filtering, three position--velocity Kalman filters, and three Vicon--gyroscope angle filters.

PID gains were tuned hierarchically on hardware---rate/attitude, rebound/aerial collective, then touchdown/landing---and frozen. Both stacks share Vicon/IMU sources and motor/PWM hardware but differ internally; PID has its own estimation, phase scheduling, ballistic prediction, cascaded attitude control, and landing feedback.

\subsection{Power, Thrust, and Cycle Normalization}

An INA260 voltage/current/power monitor supplies measurements for $P_{\mathrm{elec},t}$ via~\eqref{eq:measured_power}. Logged Vicon state and PWM commands yield $F_z/(mg)=(\sum_{i=1}^{4}F_i)/(mg)$ through~\eqref{eq:thrust_mapping}. The latest telemetry pair is stored at 200~Hz without additional filtering for both stacks.

Lowest-height points delimit cycles. Fig.~\ref{fig:power_thrust} summarizes cycle-averaged measured power and $F_z/(mg)$ for six PPO and five PID consecutive cycles; B1--B6 are successive boundaries. Mean $\pm$ standard deviation (SD) is descriptive, not independent replication. Fig.~\ref{fig:descent_velocity} normalizes apex (0\%) to maximum compression (100\%); ten complete cycles were selected consecutively from the same full record as the Fig.~\ref{fig:power_thrust} segment, with contact near minimum pre-compression velocity and $\pm1$ SD shading.

\subsection{Height, Landing, and Battery Tests}

Height tracking is evaluated at 0.8, 1.0, 1.2, and 1.4~m using 20 Vicon-detected apexes per setting.

Each control stack performs 20 landing trials per height; the next post-liftoff contact gives radial error $\sqrt{(p_x-p_x^*)^2+(p_y-p_y^*)^2}$. Battery tests run from full charge to physical failure for PPO jumping, PID jumping, and PID hovering; physical failure is loss of stable hopping or hovering followed by vehicle fall, and 3.00~V is the manufacturer reference.

\section{Results}

\subsection{Electrical Power and Rotor Thrust}

\begin{figure}[!ht]
    \centering
    \includegraphics[width=\linewidth]{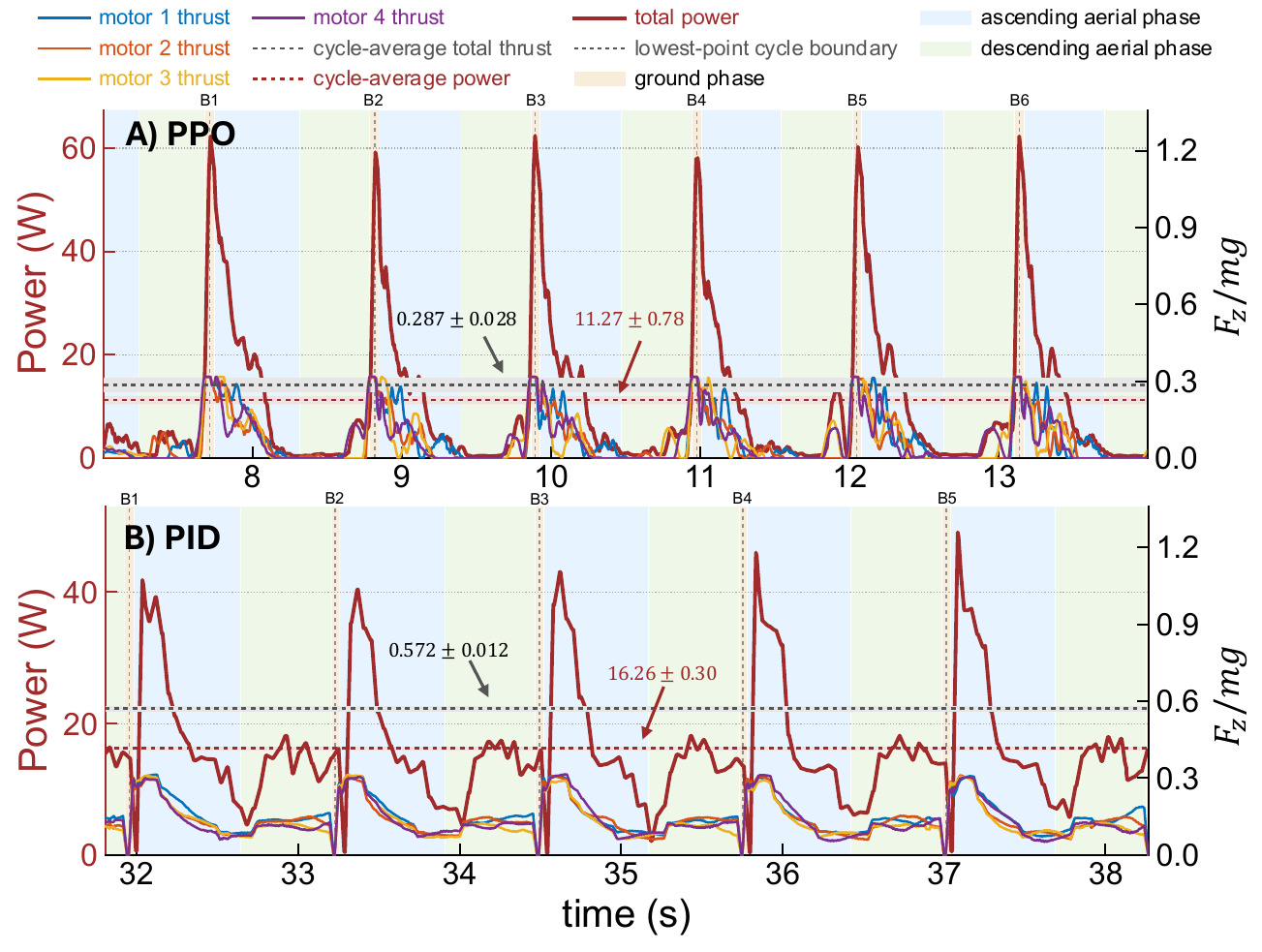}
    \caption{Measured $P_{\mathrm{elec},t}$ and $F_z/(mg)$ during one 1.2~m run per control stack. Only a representative segment of each longer hardware record is displayed. B1--B6 mark successive lowest-point cycle boundaries; annotations give descriptive mean $\pm$ SD across six PPO and five PID cycle averages, not independent hardware replicates.}
    \label{fig:power_thrust}
\end{figure}

Across consecutive 1.2-m cycles, the PPO-based control stack exhibits $11.27\pm0.78$~W and $F_z/(mg)=0.287\pm0.028$, versus $16.26\pm0.30$~W and $0.572\pm0.012$ for the tuned PID-based control stack (Fig.~\ref{fig:power_thrust}). These descriptive cycle statistics give 30.7\% lower mean measured electrical power and 49.8\% lower mean total normalized thrust. PPO retains brief takeoff pulses but uses lower aerial thrust, whereas PID maintains larger ascent/descent demand.

Bai \emph{et al.}'s model-driven controller reported a 0.28 PWM duty ratio over 1246~s \cite{bai2024agile}. Its explicit phase logic and our direct estimated-state-to-motor policy are complementary low-actuation approaches.

\subsection{Passive Ballistic Descent and Touchdown Velocity}

\begin{figure}[!ht]
    \centering
    \includegraphics[width=\linewidth]{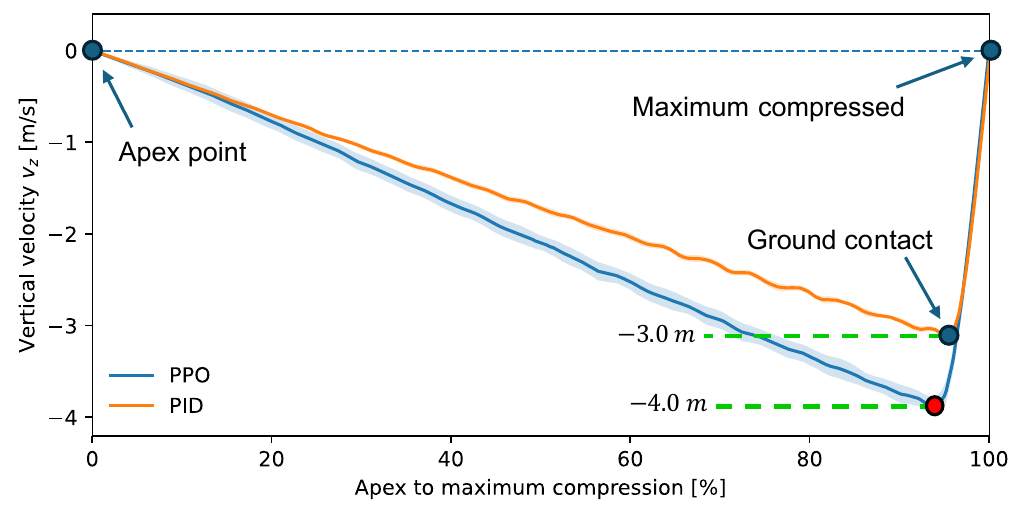}
    \caption{COM vertical velocity from apex to maximum compression; $v_z$ denotes $v_{z,t}$ with the time index omitted. The ten complete cycles were selected consecutively from the same full hardware record used for the representative segment in Fig.~\ref{fig:power_thrust}. Curves and bands show mean $\pm1$ SD; PPO/PID contact speeds are approximately $-4.0/-3.0$~m/s.}
    \label{fig:descent_velocity}
\end{figure}

From apex to contact, the PPO-based control stack reaches $-4.0$~m/s and the PID-based control stack $-3.0$~m/s (Fig.~\ref{fig:descent_velocity}). Let $\Delta h$ denote the apex-to-contact vertical drop and $v_{\mathrm{free}}$ the corresponding ballistic free-fall speed. For $\Delta h=0.90$~m,
\begin{equation}
    v_{\mathrm{free}}=\sqrt{2g\Delta h}
    =\sqrt{2(9.81)(0.90)}
    \approx4.20~\mathrm{m/s}.
\end{equation}
Fig.~\ref{fig:descent_velocity} shows mean velocity and $\pm1$ SD across $n=10$ complete cycles from the same experiment. PPO remains near the ballistic prediction, while PID applies more upward thrust. Let $E_{k,\mathrm{PPO}}$ and $E_{k,\mathrm{PID}}$ denote touchdown vertical translational kinetic energies. For equal mass,
\begin{equation}
    \frac{E_{k,\mathrm{PPO}}}{E_{k,\mathrm{PID}}}
    =\frac{(4.0)^2}{(3.0)^2}\approx1.78.
\end{equation}
Thus, the nominal vertical translational kinetic-energy ratio based on the approximate mean touchdown speeds is 1.78, corresponding to approximately 78\% greater vertical translational kinetic energy at touchdown for the PPO-based control stack.

\subsection{Commanded Apex-Height Tracking}

\begin{figure}[!ht]
    \centering
    \includegraphics[width=\linewidth]{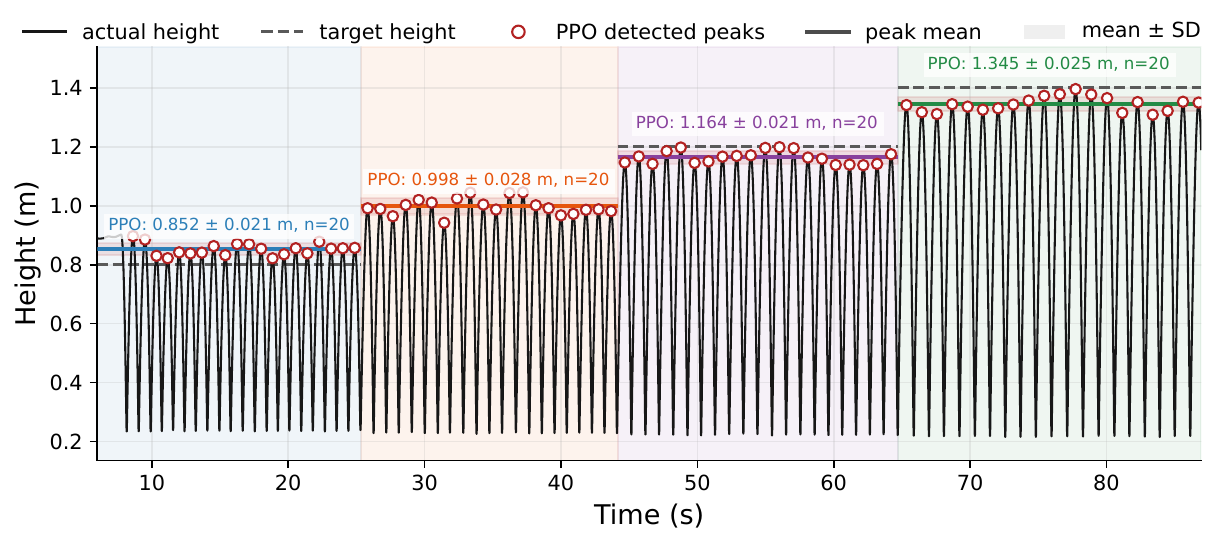}
    \caption{PPO apex-height tracking for commanded heights of 0.8, 1.0, 1.2, and 1.4~m. Red circles mark detected cycle apexes; horizontal lines and shaded bands denote the measured mean and mean $\pm$ standard deviation. The statistics are $0.852\pm0.021$, $0.998\pm0.028$, $1.164\pm0.021$, and $1.345\pm0.025$~m, respectively, with $n=20$ cycles per setting.}
    \label{fig:height_tracking}
\end{figure}

Fig.~\ref{fig:height_tracking} shows repeated hopping at all commands. At 0.8, 1.0, 1.2, and 1.4~m, means are 0.852, 0.998, 1.164, and 1.345~m; signed errors are $+0.052$, $-0.002$, $-0.036$, and $-0.055$~m, with 0.021--0.028~m standard deviations across 20 cycles. All settings remain within 0.055~m mean error.

\subsection{Target-Centered Landing Accuracy}

\begin{figure}[!ht]
    \centering
    \includegraphics[width=\linewidth]{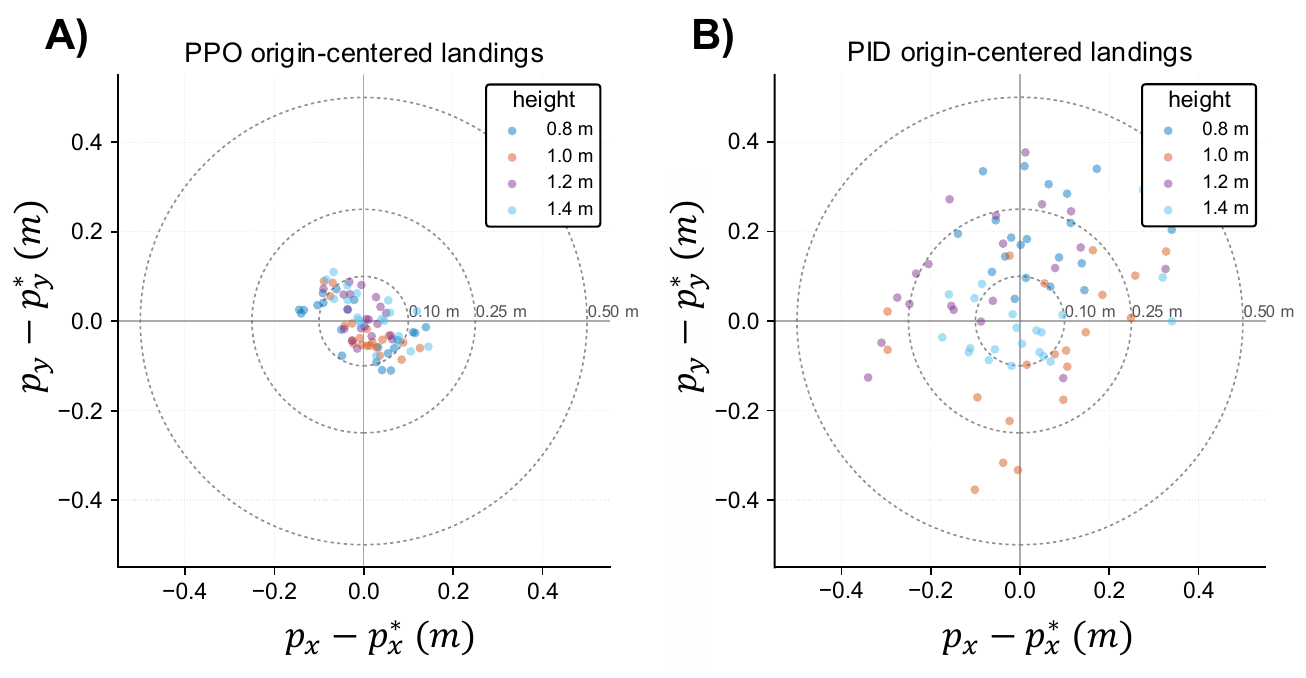}
    \caption{Target-centered horizontal landing components $p_x-p_x^*$ and $p_y-p_y^*$ at apex heights of 0.8--1.4~m. PPO is shown in the left panel and PID in the right; PPO points are more concentrated around the origin.}
    \label{fig:landing_distribution}
\end{figure}

\begin{figure}[!ht]
    \centering
    \includegraphics[width=\linewidth]{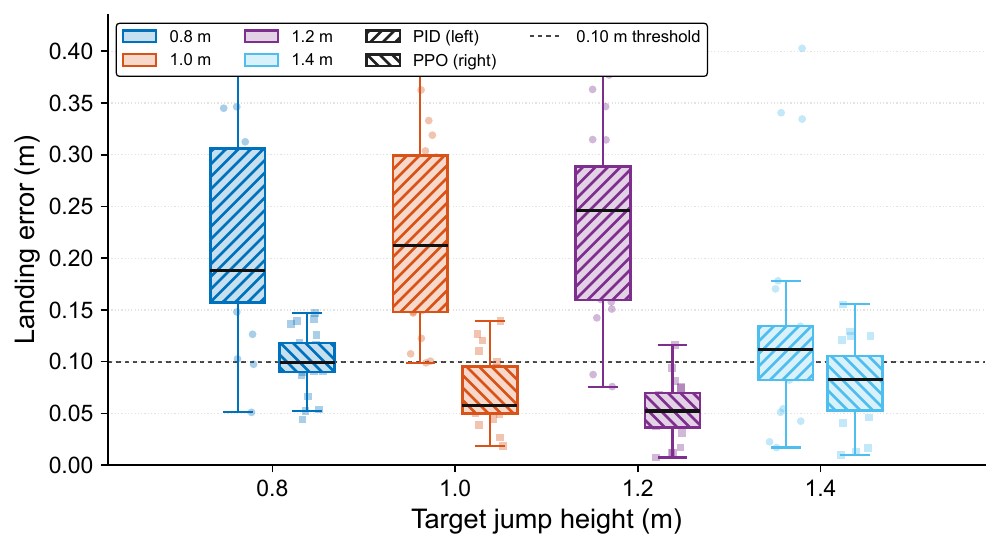}
    \caption{Paired radial landing-error box plots for 20 trials per control stack and height. PID is left and PPO is right within each pair; the dashed line marks 0.10~m.}
    \label{fig:landing_error}
\end{figure}

Fig.~\ref{fig:landing_distribution} shows a more compact PPO landing position cluster, while Fig.~\ref{fig:landing_error} pairs PID/PPO at each height. Their medians are approximately 0.19/0.10, 0.21/0.06, 0.25/0.05, and 0.11/0.08~m at 0.8, 1.0, 1.2, and 1.4~m. PPO remains near or below 0.10~m; PID has wider boxes and errors up to 0.40~m. The plots show more concentrated and repeatable PPO landings in the tested trials.

\subsection{Observed Operational Endurance and Failure Mechanisms}

\begin{figure}[!ht]
    \centering
    \includegraphics[width=\linewidth]{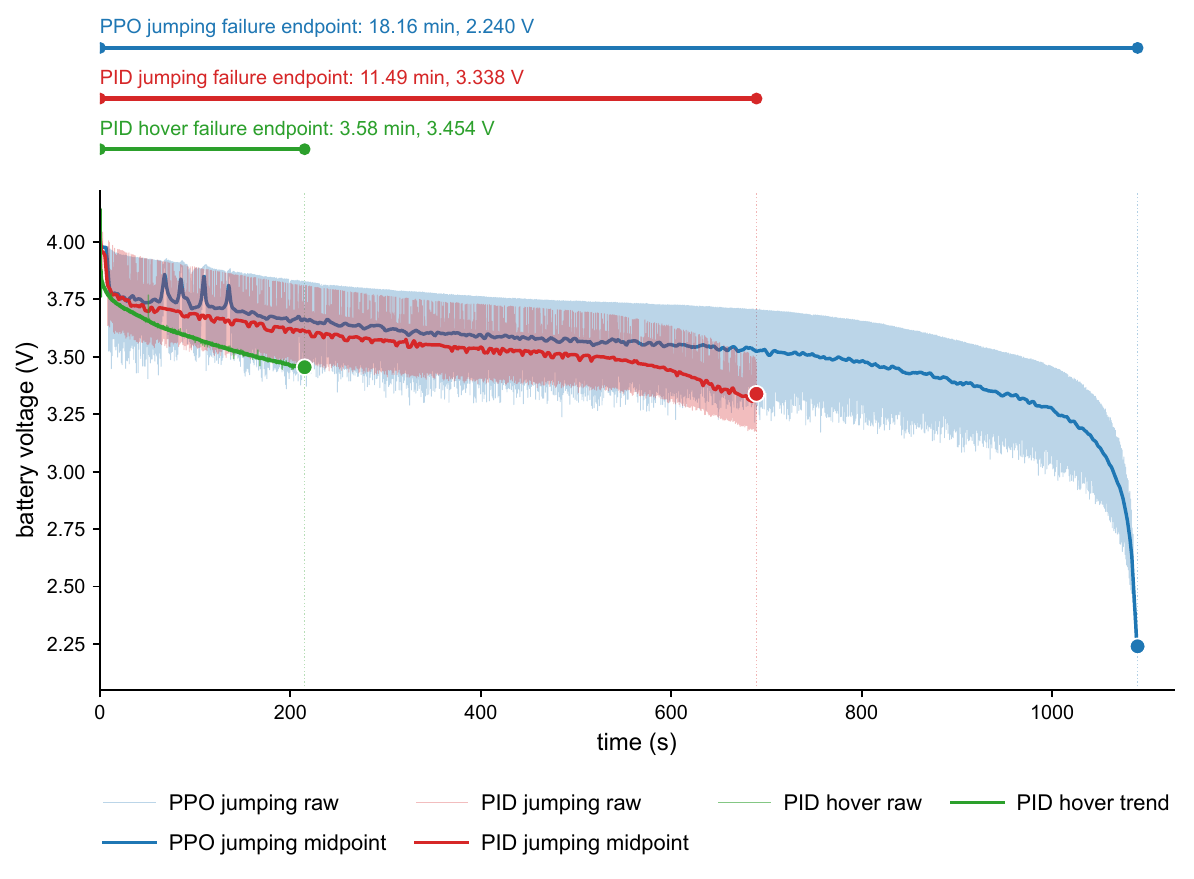}
    \caption{Single measured operational-endurance run per condition and observed physical-failure endpoint, defined as loss of stable hopping or hovering followed by vehicle fall. Midpoint curves are the pointwise midpoints between the measured upper and lower voltage envelopes. The dashed line is the manufacturer's 3.00~V discharge cut-off reference; conditions terminate at different voltages.}
    \label{fig:battery_endurance}
\end{figure}

Fig.~\ref{fig:battery_endurance} reports one endurance run per condition, all using loss of stable hopping or hovering followed by vehicle fall as the physical-failure principle. Observed durations are 18.16~min for PPO jumping, 11.49~min for PID jumping, and 3.58~min for PID hovering, with different end voltages of 2.24, 3.34, and 3.45~V. The descriptive duration ratios are 1.58 and 5.07; without repeated trials, they are not statistically generalizable battery-lifetime factors.

\subsection{Reward-Module and Estimator-Level Ablation Results}

\begin{figure}[!ht]
    \centering
    \includegraphics[width=\linewidth]{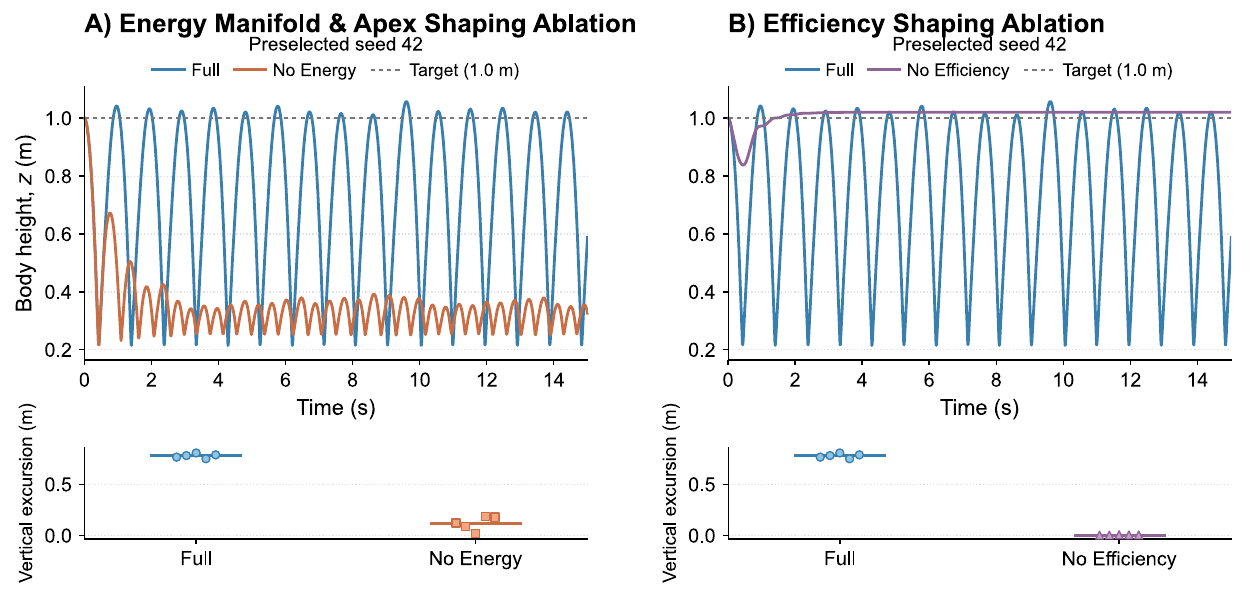}
    \caption{Energy-Manifold and Efficiency-Shaping ablations in the common vertical task. Top: unsmoothed body-height trajectories from the preselected training seed 42 for Full and the corresponding ablation. All policies start level and motionless at $(0,0,1)$~m and target $(0,0,1)$~m. Bottom: steady vertical excursion, defined independently for each training seed as $P_{95}(z)-P_{5}(z)$ over $2\leq t\leq15$~s. Markers denote seeds 42--46, and horizontal bars denote their medians; these bars are not error bars or confidence intervals. Across-seed trajectories are not averaged because phase differences between periodic trajectories would distort the apparent hopping amplitude.}
    \label{fig:reward_ablation}
\end{figure}

Fig.~\ref{fig:reward_ablation} isolates the shaping modules. Its lower metric is $\Delta z_{\mathrm{steady}}=P_{95}\{z(t):2\leq t\leq15~\mathrm{s}\}-P_{5}\{z(t):2\leq t\leq15~\mathrm{s}\}$, where $P_{95}$ and $P_5$ are the 95th and 5th percentiles. Across five descriptive training replicates, median excursion is approximately 0.781~m for Full, 0.122~m for No Energy, and 0.00356~m for No Efficiency. Removing Energy-Manifold Shaping substantially reduces excursion; removing Efficiency Shaping yields an almost stationary rotor-supported solution near the target height.

\begin{table}[!ht]
\caption{Estimator-level ablation of the electrical-power representation.}
\label{tab:power_proxy_ablation}
\centering
\scriptsize
\setlength{\tabcolsep}{3pt}
\begin{tabularx}{\columnwidth}{@{}Ycc@{}}
\toprule
Metric & Full & Power Proxy \\
\midrule
Sustained target hopping & 5/5 & 0/5 \\
Mean seed-level median apex error (m) & 0.043 & 0.639 \\
Mean seed-level median apex height (m) & 1.043 & 0.361 \\
Model-estimated energy over $0<t\leq2.5$~s (J) & 24.78 & 19.37 \\
\bottomrule
\end{tabularx}
\vspace{1pt}
\parbox{\columnwidth}{\scriptsize \emph{Note:} Five independently trained seeds (42--46) are evaluated per method using the 15-s behavioral criterion above; success counts are descriptive, with no hypothesis test. Two Power Proxy runs terminated and three formed low-height cycles. Reported energy for both policies uses the same frozen history-aware power estimator over $0<t\leq2.5$~s.}
\end{table}

\begin{figure}[!t]
    \centering
    \includegraphics[width=0.96\linewidth]{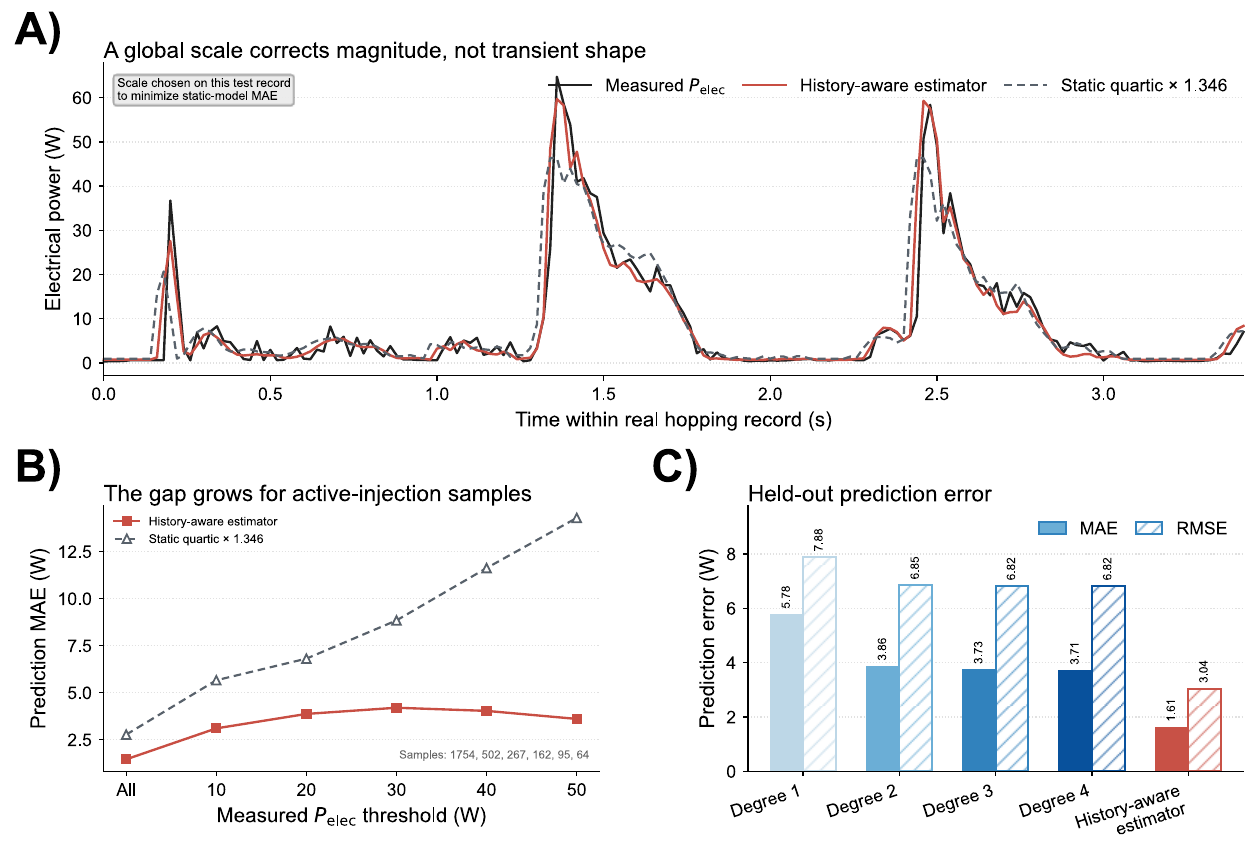}
    \caption{Prediction-level power-estimator validation. \textbf{(A)}, Independent 1.3-m record: measured $P_{\mathrm{elec},t}$, the frozen history-aware power estimator, and the static quartic estimator with $\beta_{\mathrm{oracle}}=1.35$, a deliberately optimistic, label-informed calibration not evaluated as a held-out estimate. It corrects global amplitude but not history-dependent transient shape. \textbf{(B)}, At each threshold $P_{\mathrm{th}}$, MAE is computed over samples satisfying $P_{\mathrm{elec},t}\geq P_{\mathrm{th}}$; the sample count decreases as the threshold increases. MAEs for the static quartic estimator/history-aware power estimator are 2.78/1.46~W overall and 11.63/4.04~W for $P_{\mathrm{elec},t}\geq40$~W. \textbf{(C)}, Held-out MAE/RMSE for degree-1 to degree-4 static estimators and the history-aware power estimator on 1,082 temporally correlated samples; no sample-level inference.}
    \label{fig:power_estimator_validation}
\end{figure}

Across five seeds, Full/Power Proxy sustains target hopping in 5/5 and 0/5 evaluations. Mean seed-level median apex error rises from 0.043 to 0.639~m, median apex height falls from 1.043 to 0.361~m, and estimated energy over $0<t\leq2.5$~s is 24.78/19.37~J. The lower proxy value reflects task failure, not improved energy economy. This tested-proxy behavior result does not generalize to all memoryless estimators; Fig.~\ref{fig:power_estimator_validation} separately validates static estimators at prediction level.

\addtolength{\textheight}{3pt}

\subsection{History-Aware Power Estimator Validation}

The static quartic estimator receives label-informed $\beta_{\mathrm{oracle}}=1.35$ on the independent 1.3-m record; this amplitude scale cannot restore transient shape. Its overall/high-power ($P_{\mathrm{elec},t}\geq40$~W) MAE is 2.78/11.63~W, versus 1.46/4.04~W for the history-aware power estimator.

On the separate 1,082-sample record, degree-1 to degree-4 static estimators reduce MAE from 5.78 to 3.71~W and saturate at 6.82~W RMSE; the history-aware estimator achieves 1.61~W MAE and 3.04~W RMSE (Fig.~\ref{fig:power_estimator_validation}(C)). Without sample-level inference, neither rescaling nor degree gives an equivalent prediction-level representation.

\subsection{Integrated Physical Interpretation}

\balance
Energy-Manifold Shaping promotes periodic hopping, while Efficiency Shaping penalizes power and airborne near-stationarity. Lower rotor demand and near-ballistic descent are consistent with passive-dynamics use, although spring return, recovery, and motor-off fraction are unmeasured.

\section{Conclusion}

We presented a PPO-based control stack using Energy-Manifold Shaping, Efficiency Shaping, and a history-aware power estimator. In separate runs on the same hardware, it reduced cycle-averaged measured electrical power by 30.7\% and normalized thrust by 49.8\% relative to the PID-based stack, with apex tracking, near-ballistic descent, tighter landings, and longer endurance. This supports improved passive-dynamics use but not attribution to PPO alone. Ablations linked periodic hopping to Energy-Manifold Shaping and avoidance of powered near-stationarity to Efficiency Shaping; the tested mean-matched quadratic Power Proxy failed to sustain target hopping across all five seeds.

\bibliographystyle{IEEEtran}
\begingroup
\renewcommand{\footnotesize}{\fontsize{6.9}{7.0}\selectfont}
\bibliography{IEEEexample}
\endgroup

\end{document}